\documentclass[acmtog]{acmart}

\AtBeginDocument{%
  }

\copyrightyear{2025}
\acmYear{2025}
\setcopyright{cc}
\setcctype[4.0]{by-nc-nd}
\acmConference[SIGGRAPH Conference Papers '25]{Special Interest Group on Computer Graphics and Interactive Techniques Conference Conference Papers }{August 10--14, 2025}{Vancouver, BC, Canada}
\acmBooktitle{Special Interest Group on Computer Graphics and Interactive Techniques Conference Conference Papers (SIGGRAPH Conference Papers '25), August 10--14, 2025, Vancouver, BC, Canada}
\acmDOI{10.1145/3721238.3730761}
\acmISBN{979-8-4007-1540-2/2025/08}

\acmConference[SIGGRAPH '25]{Make sure to enter the correct
  conference title from your rights confirmation email}{August 10--14,
  2025}{Vancouver, BC, Canada}

\acmSubmissionID{papers\_1476s1}

\makeatletter
\DeclareRobustCommand\onedot{\futurelet\@let@token\@onedot}
\def\@onedot{\ifx\@let@token.\else.\null\fi\xspace}

\def\eg{\emph{e.g}\onedot} 
 
\def\cf{\emph{cf}\onedot}

\makeatother

\usepackage{tikz}
\usepackage[dvipsnames]{xcolor}

\usepackage{calc}
\def\checkmark{\tikz\fill[Green, scale=0.4](0,.45) -- (.25,0) -- (1,.7) -- (.30,.30) -- cycle;} 
\def\scalecheck{\resizebox{\widthof{\checkmark}*\ratio{\widthof{x}}{\widthof{\normalsize x}}}{!}{\checkmark}}

\usepackage{hyperref}
\usepackage[linesnumbered,ruled]{algorithm2e}
\usepackage{spverbatim}
\usepackage{verbatim}
\usepackage[breakable]{tcolorbox}

\usepackage[table]{xcolor}  %
\usepackage{array}

\definecolor{myGreen}{rgb}{0.698, 0.875, 0.541}
\definecolor{myRed}{rgb}{0.984, 0.603, 0.600}

\newcommand{\prefbar}[1]{%
  \begin{tikzpicture}[
      baseline={(current bounding box.center |- 0,2pt)},
      outer sep=0,
    ]
    \def\barwidth{2.7}   %
    \def\barheight{0.31}  %

    \pgfmathsetmacro{\redFrac}{#1/100}
    \pgfmathsetmacro{\greenFrac}{1 - \redFrac}

    \pgfmathsetmacro{\oursPct}{100 - #1}

    \draw[gray] (0,0) rectangle (\barwidth,\barheight);

    \fill[myGreen] (0,0) rectangle ({\greenFrac*\barwidth}, \barheight);

    \fill[myRed] ({\greenFrac*\barwidth},0) rectangle (\barwidth,\barheight);

    \draw[
      black,
      line width=0.3pt,
      dash pattern=on 0.3pt off 0.5pt  %
    ] 
      ({0.5*\barwidth},0) -- ({0.5*\barwidth},\barheight);

    \node[
      font=\normalsize,
      anchor=west,
      inner sep=1pt
    ] 
      at (0.02*\barwidth, 0.5*\barheight)
      {\normalsize \pgfmathprintnumber[fixed,precision=1]{\oursPct}\%};

    \node[
      font=\normalsize,
      anchor=east,
      inner sep=1pt
    ]
      at (\barwidth - 0.02*\barwidth, 0.5*\barheight)
      {\normalsize \pgfmathprintnumber[fixed,precision=1]{#1}\%};
  \end{tikzpicture}%
}

\begin{document}

\title{EditDuet: A Multi-Agent System for Video Non-Linear Editing}

\author{Marcelo Sandoval-Casta\~neda}
\email{marcelo@ttic.edu}
\orcid{0000-0002-3227-9195}
\affiliation{%
  \institution{TTI-Chicago}
  \country{USA}
}

\author{Bryan Russell}
\affiliation{%
  \institution{Adobe}
  \country{USA}
}
\email{brussell@adobe.com}

\author{Josef Sivic}
\affiliation{%
  \institution{Adobe}
  \country{USA}
  \and
  \institution{Czech Institute of Informatics, Robotics and Cybernetics, Czech Technical University}
  \country{Czech Republic}
}
\email{inr03127@adobe.com}

\author{Gregory Shakhnarovich}
\affiliation{%
  \institution{TTI-Chicago}
  \country{USA}
}
\email{gregory@ttic.edu}

\author{Fabian Caba Heilbron}
\affiliation{%
 \institution{Adobe}
 \country{USA}
}
\email{caba@adobe.com}

\renewcommand{\shortauthors}{Sandoval-Casta\~neda et al.}

\begin{teaserfigure}
    \begin{center}
        \href{https://mudtriangle.com/editduet}{\textcolor{blue}{\huge{\texttt{https://mudtriangle.com/editduet}}}}
    \end{center}
\end{teaserfigure}

\begin{teaserfigure} %
  \includegraphics[width=\textwidth]{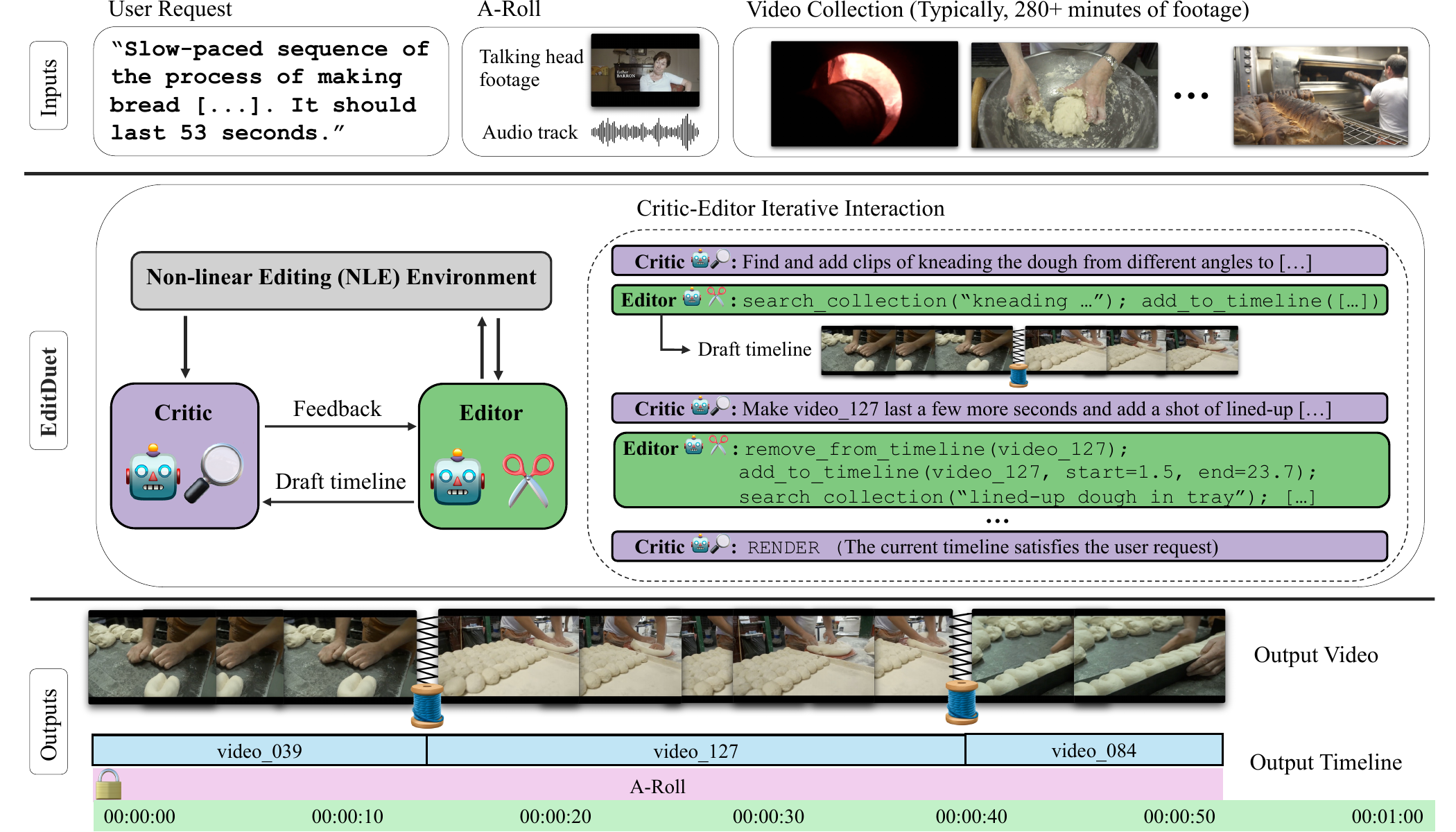}
  \caption{
  \textbf{EditDuet's summary figure}. Our system takes as input a user request, an A-roll sequence that is typically interview footage or voiceover recordings, and a real-world video collection (e.g. \href{https://app.frame.io/reviews/87ae4f06-3ac9-4cdd-91ed-cc59f0565bb3}{\textcolor{blue}{``The Ovens of Cappoquin'' from EditStock}}). We introduce two large language model (LLM) agents, an Editor and a Critic. These agents interact between themselves through feedback and the non-linear editing (NLE) timeline until the user request is considered satisfied. Once EditDuet determines that the timeline is satisfactory, this final timeline with the edits produced by our system is rendered into an output video. All thumbnails are copyrighted and belong to EditStock.}\label{fig:teaser-figure}
\end{teaserfigure}

\begin{abstract}
Automated tools for video editing and assembly have applications ranging from filmmaking and advertisement to content creation for social media.
Previous video editing work has mainly focused on either retrieval or user interfaces, leaving actual editing to the user.
In contrast, we propose to automate the core task of video editing, formulating it as sequential decision making process. Ours is a multi-agent approach.
We design an Editor agent and a Critic agent.
The Editor takes as input a collection of video clips together with natural language instructions and uses tools commonly found in video editing software to produce an edited sequence.
On the other hand, the Critic gives natural language feedback to the editor based on the produced sequence or renders it if it is satisfactory.
We introduce a learning-based approach for enabling effective communication across specialized agents to address the language-driven video editing task.
Finally, we explore an LLM-as-a-judge metric for evaluating the quality of video editing system and compare it with general human 
preference.
We evaluate our system's output video sequences qualitatively and quantitatively through a user study and find that our system vastly outperforms existing approaches in terms of coverage, time constraint satisfaction, and human preference.
\textbf{Please see our \href{https://mudtriangle.com/editduet}{\textcolor{blue}{project website}} and supplementary video for qualitative results.}
\end{abstract}


\maketitle

\section{Introduction}
\label{sec:intro}

Video non-linear editing (NLE) software, such as Adobe Premiere Pro~\cite{premiere}, Avid Media Composer~\cite{avid}, and DaVinci Resolve~\cite{davinci}, enable content creators (\eg, movie directors, ad campaigners, social media influencers) to produce, refine, and finalize compelling videos.
These applications provide tools for organizing, arranging, and compositing raw video and audio into a coherent narrative.
These tools are complex and often require advanced training\footnote{
There are many university classes that focus on this specifically, such as \href{https://catalog.registrar.ucla.edu/course/2023/filmtv154?siteYear=2023}{\textcolor{blue}{UCLA}}, \href{https://www.sps.nyu.edu/homepage/academics/courses/PUBB1-GC3456-workshop-in-video-editing.html}{\textcolor{blue}{NYU}}, and \href{https://catalogue.usc.edu/preview_course_nopop.php?catoid=16&coid=244239}{\textcolor{blue}{USC}}. 
Most well-established film programs have versions of these courses.} and experience.
A more accessible workflow might allow a user to give a high-level directive in plain English, \eg, {\em create a slow-paced establishing video lasting 30 seconds}. 
The system would then interpret and execute the directive automatically using an input collection of video clips and a suite of editing tools.

This video editing system must be able to take in the high-level user directive, break it down into individual steps, map these steps to the suite of tools in the NLE software, and use the tools to execute the directive on a collection of input video clips.
Additionally, the system should be able to incorporate any additional feedback into the current edited video, as video NLE is a fundamentally iterative process~\cite{blinkofaneye}, and be able to handle large shooting ratios, such as in reality TV and documentaries, where 100$\times$ more footage is often shot than what appears in the final edited video.
Finally, such a system should also be versatile enough to enable different modalities of co-creation, from fully automated assembly of video sequences to human-edited sequences with minimal automation.

The video NLE problem poses several major challenges. 
First, it requires understanding the video collection in terms of content (what is depicted), style (how it is depicted), and artistic and editorial style (\eg pace, motion, transitions, types of cuts, etc).
Second, it requires proficiency in using the video non-linear editing tools on the level of a video editing professional.
Third, it requires understanding of natural language instructions including the domain-specific vocabulary required for the video editing task.

We take inspiration from recent advances in large language model (LLM) systems that have enabled users to execute language tasks, including those pertaining to visual inputs, and leverage tools given an API description~\cite{gupta2023visprog,suris2023vipergpt,wu2024avatar}.
Moreover, recent work has shown how to leverage these LLMs as agents to achieve downstream tasks such as video question answering~\cite{min2024morevqa,wang2024videoagent,choudhury2025video}.
While prior efforts have addressed a variety of downstream vision-language tasks, none have been adapted to the general video NLE problem described above.

To address the above challenges, we introduce EditDuet, a multi-agent architecture for automatic video editing given a high-level user request.
We specifically tackle the task where a B-roll sequence is produced and played over an A-roll sequence. A-roll refers to voice-over or talking head interview footage, which is typically assembled first, and B-roll is complementary video that aids storytelling. This format is most common in documentaries, wedding videos, and social media content.
As this is a complex open-ended task, it is important that our agents have two properties: (1) they can engage in iterative refinement, exploring plausible solutions and determining how to improve them, and (2) each agent is specialized to a specific aspect of the solution, with a narrow set of tools focused specifically on their task.
Single agent approaches fall short on this task, resulting in incomplete or limited outputs due to their inability to keep track of multiple constraints across several steps.

Figure~\ref{fig:teaser-figure} provides an overview of our proposed system.
We propose two agents -- an editor and a critic -- based on LLMs specialized in different aspects of the video editing task.
The editor takes as input   natural language instructions, a video collection to draw sub-clips from, and the A-roll from an interview or voiceover, using existing video editing tools that interact with an NLE environment to produce the output draft timeline.
Our system can add or remove videos from a timeline, and search through large collections of videos.
The critic takes as input the draft timeline produced by the editor and the high-level user request.
It outputs feedback in natural language on the current draft timeline relative to the user request and suggest revisions for the editor.
The system is able to produce an output video that satisfies the initial user request by iterating over the critic and the editor, which progressively improves the existing timeline until the critic is satisfied and calls the renderer.

In a multi-agent setup, determining the messages that elicit desired agent behaviors is crucial.
To enable efficient communication across agents, we propose an exploration-based approach to generate synthetic demonstrations  as in-context examples for multi-agent interactions.
This way, our Editor agent learns from in-context examples the type of feedback provided by the Critic agent and how to satisfy it, while the Critic agent learns what feedback should look like to achieve a timeline that aligns with the user's request.

Finally, given the creative nature of the task, there is a large number of valid solutions for any potential user request, making it hard to collect ground-truth data.
Although studies that capture user preferences reliably compare output quality, they are costly and difficult to scale.
To address this challenge, we propose an LLM-as-a-judge~\cite{zhen2023judging} approach that automatically evaluates non-linear editing timelines by analyzing structure, relevance, aesthetic coherence, and pacing.
We show that this automatic judgment closely correlates with users' preferences, making it a valuable signal when automatically evaluating systems in this task.
Our experiments in real-world NLE projects (from \href{https://editstock.com/collections/stock-film-dailies}{\textcolor{blue}{EditStock}}) show that our method vastly outperforms existing approaches in failure rate, time constraint satisfaction, and human/automatic preference.

\section{Related Work}
\label{sec:related-work}

\subsection{AI-assisted Video Non-linear Editing}
There has been growing interest in using ML and AI for video non-linear editing (NLE).
At its core, video NLE involves selecting, arranging, trimming, and composing raw footage to assemble sequences in an NLE timeline.
Given the complexity of the task, a key question is how to formulate the video NLE task and make it suitable for ML models.
\citet{leake2017computational} formulates video NLE as a sequence modeling task, where one view is selected from time-aligned multi-cam videos based on cinematographic idioms.
\citet{xiong2022transcript} frame the video NLE task as a text-to-video alignment problem, where the goal is to retrieve a sequence of video clips from a collection that best matches a given speech transcript.
More recently, \citet{pardo2024generative} proposed Timeline Assembler, a method that takes a video collection and a user directive as input to manipulate an existing video editing timeline.
While earlier approaches focus on constrained scenarios, our method works with extensive raw footage, supports broader user input, and formulates the video NLE task as an iterative process of building a timeline using typical NLE tools like search, trimming, and adding/removing clips.

\subsection{Video Editing Tools}
Several point technologies have been proposed to tackle specific tasks in video editing. 
These include alignment of multi-camera videos~\cite{sun2023eventfulness,wang2014videosnapping,arev2014automatic}, predicting cuts in interviews ~\cite{berthouzoz2012tools}, retargeting musical performances ~\cite{lee2022popstage}, and gaze-based editing~\cite{jain2015gaze}.
Other methods predict transitions between fixed clips~\cite{shen2022autotransition,guhan2025v} or learn editing patterns~\cite{argaw2022anatomy}.
Several other works~\cite{wang2024lave,huber2019b,wang2024podreels,huh2023avscript,tilekbay2024expressedit} develop new user interfaces for interactive human-AI video editing.
Our method builds on these tools, starting with visual search and timeline actions as a foundation to integrate and expand to more capabilities.

\subsection{Iterative Refinement for LLMs}
Iterative refinement methods for LLMs often revolve around synthetically generated feedback ~\cite{madaan2024self, huang2024blenderalchemy}, scoring-based optimization ~\cite{romera2024mathematical,chiquier2024evolving}, or a mix~\cite{du2024blenderllm} to gradually improve the output quality. 
Inspired by these approaches, we iteratively improve our Editor's agent.
In contrast to single-agent refinement, we propose a dual-agent framework where an Editor handle low-level video NLE operations, and a Critic observes draft timelines produced by the editor. The Critic iteratively provides feedback in natural language to the Editor, leading to video NLE outputs with higher quality.
Note that our Critic offers natural language guidance to the Editor during inference, different from reinforcement learning actor-critic methods~\cite{konda1999actor,haarnoja2018soft}.

\subsection{Agentic Workflows Using LLMs}
LLM-based agents are now popular for automating tasks in software engineering, 3D modeling, and more~\cite{yang2024swe, hu2024scenecraft, huang2024crispr}. 
Function calling has opened many potential applications for LLMs. This has proven  useful for tasks in computer vision too, even using models that do not natively process visual input~\cite{suris2023vipergpt, gupta2023visprog, wang2024videoagent, min2024morevqa}. 
There are multiple approaches to effectively implement LLM agents, such as ReAct~\cite{yao2022react} and Husky~\cite{kim2024husky}. 
However, these approaches typically rely on high quality human-made examples of sequences of actions and queries for in-context learning.
Recently, BAGEL~\cite{murty2024bagel} proposed a method enabling LLM agents for web navigation to generate examples via exploration and self-labeling.
Unlike BAGEL's single-agent exploration, our approach enables multi-agent collaboration by finding examples that benefit both agents' communication.
We also incorporate self-reflection during self-supervised environment exploration to find high-quality demonstrations faster.

\subsection{LLM-based Evaluation}
Recently, some works~\cite{kim2024prometheus,lee2024prometheusvision,zhang2023gpt4visiongeneralistevaluatorvisionlanguage} have shown that LLM judgments of quality for other language models highly correlates with human preferences.
In particular, \citet{lee2024prometheusvision} showed that a vision-language model (VLM) can follow a detailed rubric when scoring image captions.
They show that this is strongest using closed-source very large models, in particular GPT-4V~\cite{gpt4vcontribution}, which had the highest correlation with human judgments.
Other works successfully leverage closed source VLM judgments of quality for evaluation of image generation~\cite{wu2024gpt} and 3D generation~\cite{lin2024evaluating}.

Building on these findings, we propose a video NLE judge that uses a VLM to evaluate non-linear editing timelines, and demonstrate that such automatic judgments correlate with human preferences.

\section{Approach}
\label{sec:approach}

\begin{figure*}[]
  \includegraphics[width=\textwidth]{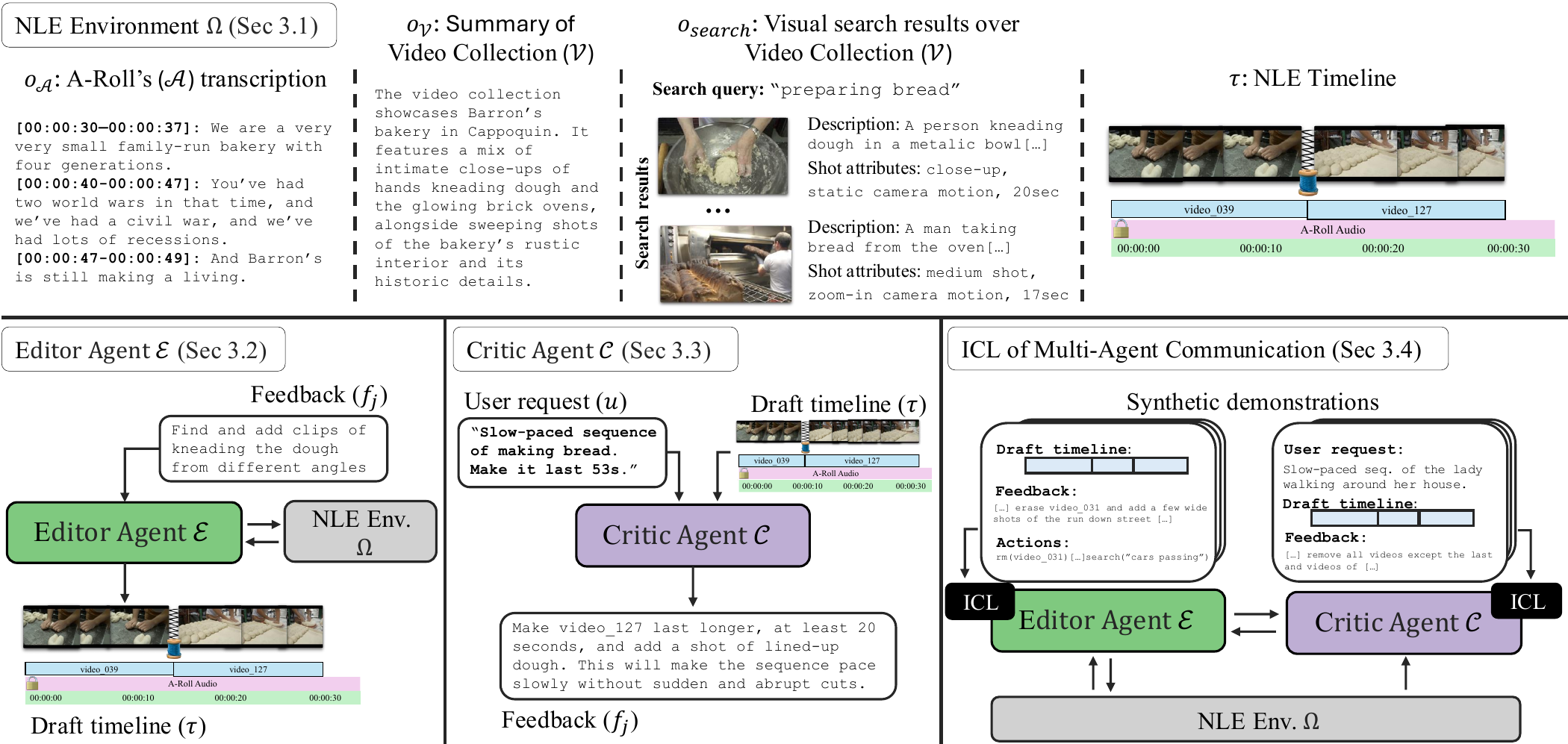}
  \caption{\textbf{Overview of our multi-agent non-linear editing framework.} The NLE environment $\Omega$ (Sec. \ref{sec:approach-env}) exposes four observations to the agents: an A-roll transcription $o_\mathcal{A}$, a summary of the video collection $o_\mathcal{V}$, visual search results $o_{search}$, and the NLE timeline $\tau$. 
  The Editor agent $\mathcal{E}$ (Sec. \ref{sec:approach-editor}) modifies $\tau$ based on feedback $f_j$ from the Critic agent $\mathcal{C}$.
 The Critic verifies if the timeline satisfies user request $u$, then either suggests edits or finalizes and renders the video (Sec.~\ref{sec:approach-critic}). Finally, we propose a two-stage process for In-Context Learning (ICL) of multi-agent communication (Sec. \ref{sec:approach-communication}), using synthetic demonstrations.
 All thumbnails are copyrighted and belong to EditStock.}
  \label{fig:system}
\end{figure*}

We propose a multi-agent system based on iterative refinement whose objective is to compose a non-linear editing (NLE) timeline within an environment, given a collection of videos, the A-roll that will play over the resulting timeline, and a user query in natural language.
We formulate this task as an iterative process where two agents -- Editor and Critic -- collaborate within an NLE environment, sequentially taking actions and refining timelines based on mutual feedback and observed outcomes.
Figure~\ref{fig:system} provides an overview of our method's components.
The NLE environment (Sec. \ref{sec:approach-env}) encapsulates the video collection and A-roll into four main observations accessible to the agents: the A-roll transcription, a summary of the video collection, a search engine that outputs visual results from text queries, and the NLE timeline itself.
The Editor agent (Sec. \ref{sec:approach-editor}) modifies the timeline based on feedback instructions from the Critic agent. 
The Critic agent (Sec. \ref{sec:approach-critic}) checks if the resulting timeline meets the user query and outputs further feedback or decides to finalize the process and render the output video.
We also propose a method for enabling effective communication between the Editor and Critic (Sec. \ref{sec:approach-communication}).
Finally, we design a VLM-based judge \cite{kim2024prometheus,lee2024prometheusvision,zhang2023gpt4visiongeneralistevaluatorvisionlanguage} to enable automatic evaluation of NLE timelines.
We provide prompts for instantiating all agents described in this section in the \textit{supplementary material}.

\subsection{Non-Linear Editing Environment}\label{sec:approach-env}
Our system is equipped with an NLE environment.
We define the environment $\Omega \left( \mathcal{A}, \mathcal{V},  \tau \right)$, where $\mathcal{A}$ is the A-roll for the documentary we are editing, which is typically a talking-head interview video, $\mathcal{V}$ is a video collection, and $\tau$ is a non-linear editing (NLE) timeline.

The video collection $\mathcal{V}$ is the set of all available videos in a user project.
Each video is segmented hierarchically into shorter video sub-clips using TW-FINCH clustering~\cite{tw-finch}.
This segmentation enables our system to return video segments during search instead of full videos, which is useful when searching through longer video files.
Clusters shorter than one second are discarded. 
Each remaining cluster has the following attributes: start time and duration in seconds, a description of the video segment's content, cinematographic shot type, and camera motion classification.
The description is produced using Llava-NeXt~\cite{zhang2024llavanext-video} to generate captions for each video segment, with exactly 16 frames per cluster.
Cinematographic shot type is one of: extreme close-up, close-up, medium, full, and long shot.
Shot types were labeled using a MobileNet V3 classifier~\cite{howard2019searching} trained on the MovieShots dataset~\cite{rao2020unified}.
Camera motion classification is one of: static, zoom in, vertical static/moving, aerial, travelling in/out, handheld, panoramic and panoramic lateral.
These camera motion types were obtained through a long short-term memory (LSTM) classifier~\cite{graves2013generating} using Single Shot MultiBox Detector (SSD) features~\cite{liu2016ssd} trained on the Movie Shot Classification dataset~\cite{petrogianni2022film}.
Additionally, each video collection is summarized in a paragraph using Llama3.1-70B-Instruct~\cite{dubey2024llama3herdmodels}, which is prompted with the top-level segment descriptions from the cluster hierarchy of each video.

The timeline $\tau$ is an ordered sequence of video sub-clips to be rendered.
In our problem setup, we focus on editing the B-roll timeline, and assume the A-roll is fixed.
Typically, this timeline starts empty, and is progressively populated by video sub-clips from $\mathcal{V}$ as our agents take actions in the environment.
Optionally, this timeline may be initialized with random sub-clips from the video collection or from a timeline obtained from a previous run.

Given the scale of the projects we tackle, with hundreds of minutes of recorded footage in the video collection $\mathcal{V}$, we also rely on a search engine for our agents to interact with the video collection.
Instead of providing the full video collection, the agents can explore the corpus through search tools, which we limit to show up to five results at a time.
Including the full video collection as part of our agents' context is often infeasible due to computation and model limitations given the video collection size.
Moreover, if the search results are not satisfactory, the agent may perform additional search steps with different queries (\cf, first green box in Figure~\ref{fig:qualitative-figure}).
Additionally, this setup allows us to model the task closer to how a human would interact with an NLE environment, where they have access to a search panel instead of looking at all videos at once.

An observation of the environment $o \sim \Omega$ is then a 4-tuple
$o = \left( o_\mathcal{A}, o_\mathcal{V}, o_\texttt{search}, \tau \right) \sim \Omega \left( \mathcal{A}, \mathcal{V}, \tau \right)$,
where $o_\mathcal{A}$ is the text transcription from the audio in the A-roll, $o_\mathcal{V}$ is the static video collection summary, $o_\texttt{search}$ is a list of video sub-clips given a search query, and $\tau$ is the current timeline state.
 
\subsection{Editor Agent}\label{sec:approach-editor}
The goal of our Editor agent $\mathcal{E}$ is to modify the timeline based on  feedback from the Critic agent.
To do so, it uses functions representing typical actions in a video NLE environment (described below). Actions are sampled one at a time by generating a function call and its corresponding parameters from the LLM backbone of our agent, given an observation of the environment, its previous action history, and feedback from the Critic agent. 
Formally, $a^\mathcal{E}_i \sim \mathcal{E} \left( o_i, h^\mathcal{E}_{i - 1}, f_j \right)$,
where $a^\mathcal{E}_i$ is the action taken by the Editor agent at Editor time step $i$, $o_i$ is an environment observation at editor time step $i$, $h^\mathcal{E}_{i - 1}$ is the observation-action history of our agent, and $f_j$ is the latest feedback provided by the Critic agent at Critic time step $j$.
History objects are of the form $h_{i} = (o_0, a_0, o_1, a_1, \ldots, o_{i}, a_{i})$ where $o_i$ is the observation of the environment at time step $i$ and $a_i$ is the action taken after observing the environment at time step $i$.

The Editor agent has access to the following functions:%

{\textbf{\texttt{search\_collection:}}}
this tool takes in a query string $s$ from the Editor and returns the five video segments with highest CLIP similarity~\cite{radford2021learning} to $s$. This is the only function in our setup that directly receives frames' pixels as input.
In the case of overlapping candidate segments,
if the longer sub-clip has at least $90\%$ similarity score of the shorter sub-clip, the shorter sub-clip is dropped.
This tool directly populates $o_\texttt{search}$, which starts empty.

{\textbf{\texttt{add\_to\_timeline:}}}
this operation takes a video $v\in\mathcal{V}$, an index $k$, a start time $t_s$ in seconds, and an end time $t_e$ in seconds, and inserts the corresponding sub-clip of the video that starts at $t_s$ and ends at $t_e$ at index $k$ of the timeline.

{\textbf{\texttt{remove\_from\_timeline:}}} 
it takes an index $k$ and removes the video at that position in the timeline.

{\textbf{\texttt{switch\_clip\_positions:}}}
it takes two indices, $k$ and $l$, and swaps the videos at each index in the timeline.

{\textbf{\texttt{move\_clip:}}} 
it takes a source index $k$, and a target index $l$, and moves the video at position $k$ of the timeline to position $l$.

{\textbf{\texttt{DONE:}}} 
it takes no parameters and signals that the feedback received has been satisfied, shifting control back to the Critic.
 
We use Llama3.1-8B-Instruct~\cite{dubey2024llama3herdmodels} as the backbone for the Editor Agent.
To mitigate broken outputs or invalid function calling, we use structured generation~\cite{willard2023efficient}.

\subsection{Critic Agent}\label{sec:approach-critic}
Our Critic agent $\mathcal{C}$ receives a timeline and a user query, and its goal is to decide whether the given timeline satisfies the query or, if not, suggest a way to improve it.
Formally,
$a^\mathcal{C}_j \sim \mathcal{C} \left( o^\tau_j, h^\mathcal{C}_j, u \right)$,
where $a^\mathcal{C}_j$ is the action taken by the Critic agent at critic time step $j$, $o^\tau_j$ is an observation of only the timeline, $h^\mathcal{C}_j$ is the critic's history, and $u$ is the user request.

The Critic agent takes one of two actions:

\textbf{\texttt{give\_feedback:}}
it receives the string $f_j$ from the Critic that will be output as feedback for the Editor agent, starting a new iteration of the Editor's process.

\textbf{\texttt{RENDER:}}
signals that the video is ready to be rendered.

In this paper, we use the LLM Llama3.1-8B-Instruct with structured generation as the Critic agent, the same backbone as our Editor Agent.
We do not use a VLM Critic due to the number of tokens required to fit multiple iterations of the resulting timeline as a video, which greatly exceeds current open source VLM context sizes.

\subsection{In-Context Learning of Multi-Agent Communication}\label{sec:approach-communication}
Early in our experimentation, we found that the Editor often failed at extracting meaningful information from the Critic's feedback, and the Critic would similarly fail at providing actionable feedback, or hallucinating actions that cannot be performed (\eg, ``shoot a new video of the scene from a different angle'').
See the \textit{supplementary material} for detailed examples.
To mitigate this issue, we propose a test-time self-supervised exploration process to find synthetic in-context examples given the NLE environment and input video collection.
This process is beneficial for the Editor to better interpret messages and for the Critic to improve its feedback, while allowing both agents to see an excerpt of the video collection through these examples.
While it would be possible to handcraft the kinds of examples our method provides, it is a difficult task: the space of possible action sequences to obtain one specific timeline is infinite and requires step-by-step decision-making. It would also require multiple demonstrations for a given video collection, at which point the desired documentary might as well be edited by hand.

Our method is based on in-context learning~\cite{dong2022survey} and does not involve any fine-tuning of the model parameters. We use a two-stage exploration to find in-context demonstrations to include in the prompts for our Editor and Critic agents. The first stage collects demonstrations of tool usage and feedback-following for the Editor, while the second stage finds examples for the Critic to guide the Editor in improving B-roll sequences.

To find Editor demonstrations, we define four auxiliary agents:

    \textbf{Editor Explorer:} explores the environment by taking actions that modify a timeline.
    
    \textbf{Editor Labeler:} predicts the natural language feedback that could have elicited the timeline changes produced by the Editor Explorer's observation-action history.
    
    \textbf{Editor Scorer:} scores the Editor Explorer's observation-action history and the Editor Labeler's corresponding feedback on a scale of 1 to 5 based on alignment and efficiency.
    
    \textbf{Self-Reflecting Editor:} analyzes high-scoring examples and attempts to improve them.

We run the Editor Explorer starting from a randomly initialized timeline until it takes the action \texttt{DONE}. Then, to simulate the Critic's feedback that yielded the exploration taken by the Editor Explorer, the Editor Labeler generates text feedback corresponding to the observation-action history of the Editor Explorer, \eg, ``currently the provided timeline is lacking [...] change the existing footage to a montage [...]''. If the Editor Scorer assigns a score of 3 or lower for the Editor Labeler's feedback of the Editor Explorer's observation-action history, the example is discarded. If the score is 4, the Self-Reflecting Editor refines the actions for re-scoring. This refinement usually takes the form of removing redundant actions or selecting different and more appropriate footage. Only pairs scoring 5 are kept as synthetic demonstrations. This process repeats until we collect five high-quality demonstrations for the Editor.

To find Critic demonstrations, we define three auxiliary agents:

    \textbf{Critic Explorer:} explores the environment by interacting with the Editor Agent that modifies a timeline initialized from drawing video sub-clips at random from the video collection.
    
    \textbf{Critic Labeler:} predicts a user request that may have elicited the feedback history produced by the Critic Explorer.
    
    \textbf{Critic Scorer:} scores a user request-timeline pair (from 1 to 5) based on how well the timeline satisfies the request.

We run EditDuet with the Editor Agent (using the in-context demonstrations found in the first stage) and the Critic Explorer until the Critic Explorer takes the action \texttt{RENDER}. The Critic Labeler then generates a user request (\eg, ``tight close-ups of an old woman kneading dough [...]''). Then, the Critic Scorer assigns this demonstration a score based on the Critic Labeler's user request and the final timeline produced by this process. Only examples scoring 5 are kept as synthetic demonstrations for the Critic Agent. This process continues until we collect five high-quality examples.

\subsection{Automatic Non-Linear Editing Evaluation}\label{sec:approach-aieval}

Evaluating non-linear editing (NLE) is inherently challenging due to the difficulty of defining reliable ground-truth data and the existence of multiple equally valid editing solutions.
While human preference studies remain the most reliable source of evaluation, they are expensive and hard to scale. 
To address these issues, we propose an automated framework based on a Vision-Language Model (VLM) that serves as an automatic NLE judge.

Concretely, we define the automatic NLE judge $\mathcal{J}$ as a function that takes the user request and two non-linear editing timelines $\tau_1$ and $\tau_2$ and outputs its preferred choice $\mathcal{J}(\tau_1,\tau_2) \in \{\tau_1,\tau_2\}$.
We argue that a sufficiently capable VLM, prompted with the right information about each timeline's structure and content, can approximate human preferences for non-linear editing quality.

In practice, we leverage GPT-4o as our VLM. 
We prompt the model with a grid of keyframes from the midpoint of each sub-clip alongside each sub-clip's duration. 
This setup enables the judge to visualize the overall structure of the timeline, assess the relevance of the selected B-roll, gauge the pacing among sub-clips, and capture overall aesthetic coherence. 
By analyzing the keyframe grids for $\tau_1$ and $\tau_2$, the judge can generate its preference, which can be seen as a proxy for human evaluation of editing quality.

Beyond pairwise comparisons, we use $\mathcal{J}$ to evaluate different editing methods at scale. 
Given methods $M_1$ and $M_2$, we define the $\mathrm{PreferenceRate}$ of $M_1$ over $M_2$ as the fraction of episodes a timeline from method $M_1$ is preferred by $\mathcal{J}$ over a timeline from method $M_2$.
Since it uses our automatic NLE judge, this metric provides a scalable strategy to quantify NLE performance.

\subsubsection{NLE judge correlation with human preference.} To validate the reliability of our automatic NLE judge, we collected human preferences on pairs of timelines generated by a diverse set of methods.
We did so by collecting the judgments of 35 people through an online survey. Each person was presented with ten pairs of edited videos for the same B-roll, sampled at random from our results, and asked to choose their preferred edited video without information on the method that produced it.
For each pair of videos, we calculate agreement for our automatic method by comparing against the majority-voted video for each pair in the user study.
We observe an 80.6\% agreement between the judge's choices and human preferences, compared to 78.7\% agreement among different human evaluators themselves. Furthermore, the Prevalence-Adjusted Bias-Adjusted Kappa (PABAK)~\cite{byrt1993bias}--chosen to address the unbalanced distribution of ``preferred'' choices--was 0.61 for the judge–human comparison, closely matching the 0.57 obtained among humans.
These results suggest that the NLE judge can be as reliable as a human in evaluating non-linear editing quality.

\section{Results}
\label{sec:results}

\subsection{Evaluation Dataset}
To evaluate our method, we use data from EditStock~\cite{editstock}.
EditStock is a website that compiles the raw footage and final cuts from several filmmaking projects across genres and styles.
At the time of writing, there are fifty six projects available, spanning fiction films, documentaries, sports videos, and others.

We focus on five documentary titles: ``The Scramble King'', ``The Rock Climber'', ``The Ovens of Cappoquin'', ``Shores of this Bay'', and ``Built by Life''.
We focus on sequences that contain B-roll for our task, and manually annotate each one of these documentaries with high-level prompts that describe the style and semantic content of each B-roll sequence to serve as user request for the editing systems.
For example, ``tight close-ups of an old woman kneading dough, interleaved with close-ups of her face while baking. End with fast-paced wide shots of the oven setup process. It lasts for 18 seconds.'' is an annotation for the documentary ``The Ovens of Cappoquin''.
All our annotations follow the same structure and content.
Each sequence of B-roll is up to 1.25 minutes in length, and consists of up to 27 sub-clips.
In total, this is 1458 minutes of video across five documentaries to be edited down to 21.5 minutes of final cut videos.

\begin{table*}[]
\begin{tabular}{lccccccccc}
\hline
\multicolumn{1}{c}{Method} & Func. Call. & Trim. & Multi-Agent & Expl. & Fail. Rate $\downarrow$ & Cov. $\uparrow$ & Reps. $\downarrow$ & Human Preference & Auto. Preference \\
\hline
\textbf{Baselines} \\
\hline
T2V &  &  &  &  & 0.0\% & 92.6\% & 2.696 & \prefbar{13.1} & \prefbar{22.2}\\
VisProg & \scalecheck &  &  &  & 34.8\% & 44.8\% & 0.783 & \prefbar{16.1} & \prefbar{18.5} \\
BAGEL & \scalecheck & \scalecheck &  & \scalecheck & 14.3\% & 73.5\% & 0.214 & \prefbar{18.2} &  \prefbar{11.5}\\
\hline
\textbf{Ablations} \\
\hline
Editor Only & \scalecheck & \scalecheck &  &  & 23.8\% & 68.5\% & 0.217 & \prefbar{14.3} &  \prefbar{22.2}\\
Editor Critic & \scalecheck & \scalecheck & \scalecheck &  & 19.5\% & 82.7\% & 0.257 & \prefbar{35.1} & \prefbar{29.4}\\
\hline
EditDuet & \scalecheck & \scalecheck & \scalecheck & \scalecheck & 8.2\% & 89.8\% & 0.174 & N/A & N/A \\
\hline
\end{tabular}
\caption{\textbf{Comparison between our method, EditDuet, and other baselines and ablations.} Function calling (Func. Call.) refers to calling functions. Trimming (Trim.) refers to selecting  sub-clips beyond the input segments. Exploration (Expl.) refers to whether a system has an exploration stage. Failure rate (Fail. Rate) is the percentage of runs where a system fails to render a video, time coverage (Cov.) is the ratio between the desired time sequence and the output video, and repetitions (Reps.) is the average number of repeated sub-clips in each output video. Human preference and automatic preference are described in Sec. ~\ref{sec:approach-aieval}. The \textcolor{myGreen}{green bar} indicates preference for our approach. EditDuet yields fewest repetitions, highest preference, and ranks top-2 in failure and coverage.  
}\label{tab:results-main}
\end{table*}

\subsection{Evaluation Criteria}
We evaluate the systems using PreferenceRate described in Section~\ref{sec:approach-aieval} and other metrics beyond editing quality.
For each of the methods studied, we provide the average time coverage, average number of repetitions per sequence, and failure rate.
We define time coverage as the ratio $TC \left( d, \hat{d} \right) = \min\left(d, \hat{d}\right)/\max\left(d, \hat{d}\right)$,
where $d$ is the duration in seconds of the ground truth sequence corresponding to a user request, and $\hat{d}$ is the duration in seconds of the video produced by the system.
For number of repetitions per sequence, we consider a repetition a pair of sub-clips within the same output sequence where there is at least $80\%$ overlap. If a given clip appears more than $n > 2$ times, this is counted as $n - 1$ repetitions.
Finally, we include failure rate as our agents and competing methods are not all guaranteed to always output valid actions and timelines. Examples failures include invalid file names and out-of-bounds indices.

\subsection{Baselines and Competing Methods}
We compare our method with two non-agentic baselines, Transcript2Video (T2V)~\cite{xiong2022transcript} and VisProg~\cite{gupta2023visprog}, and an agentic one, BAGEL~\cite{murty2024bagel}.

T2V is a retrieval-based method specifically for finding footage related to some narration among a large collection of candidates.
As such, it is well-suited for our task of constructing B-roll for interview-based documentaries.
This is the only method not based on LLMs that we compare to.
Given that the implementation of the original method is not available, we adapt it by using CLIP similarity-based search to select relevant footage.
Additionally, since our videos do not come pre-sliced in relevant clips, we use video segments extracted via TW-FINCH clustering as outlined in Section~\ref{sec:approach-env}.
This method does not use any of our sub-clip labels, as it is entirely retrieval-based using CLIP.

VisProg is a method for extracting executable programs for vision tasks given a fixed set of functions.
We extend it for our task by having our agents' set of actions as functions in VisProg's API, and allowing VisProg's programs to be executed within our editing environment.
For simplicity, we use video segments extracted via TW-FINCH and allow VisProg to directly operate with search result objects
in the function calls.

We run BAGEL for a single Editor agent, augmenting it with demonstrations drawn from exploration. We do so in each video collection separately, as in our method.

Finally, we also perform ablations of our system and include the comparisons as results. We start from a single Editor agent, as described in the previous section. We then compare our final system with an Editor-Critic setup without any exploration, to demonstrate the effect in performance and quality of this step in the process.

\subsection{Results and Analysis}
Table \ref{tab:results-main} compares several non-linear B-roll editing methods. Our final approach ("EditDuet") has the lowest failure rate (8.2\%) among LLM-based methods. It also achieves near-ideal time coverage (89.8\%) while minimizing repeated footage (0.174). Tanscript2Video~\cite{xiong2022transcript} covers target durations well but often repeats the same sub-clips, resulting in a lower preference rate (13.1\% human, 22.2\% automatic). VisProg \cite{gupta2023visprog} exhibits a high failure rate (34.8\%) and poor coverage (44.8\%), partly due to its inability to refine actions iteratively. BAGEL \cite{murty2024bagel}, while it includes an exploration phase, does not produce a high preference rate (18.2\% human, 11.5\% automatic) because it focuses on how to use the editor's tools rather than on actions that yield high quality edits. Ablations confirm that both multi-agent collaboration and our self-supervised exploration phase are critical. Without them, the system fails more often, covers less time, or repeats clips. Overall, our Editor-Critic setup outperforms the competing methods in both quantitative metrics and automatic preference judgments.

\subsection{Qualitative results}
\begin{figure*}
  \includegraphics[width=\textwidth]{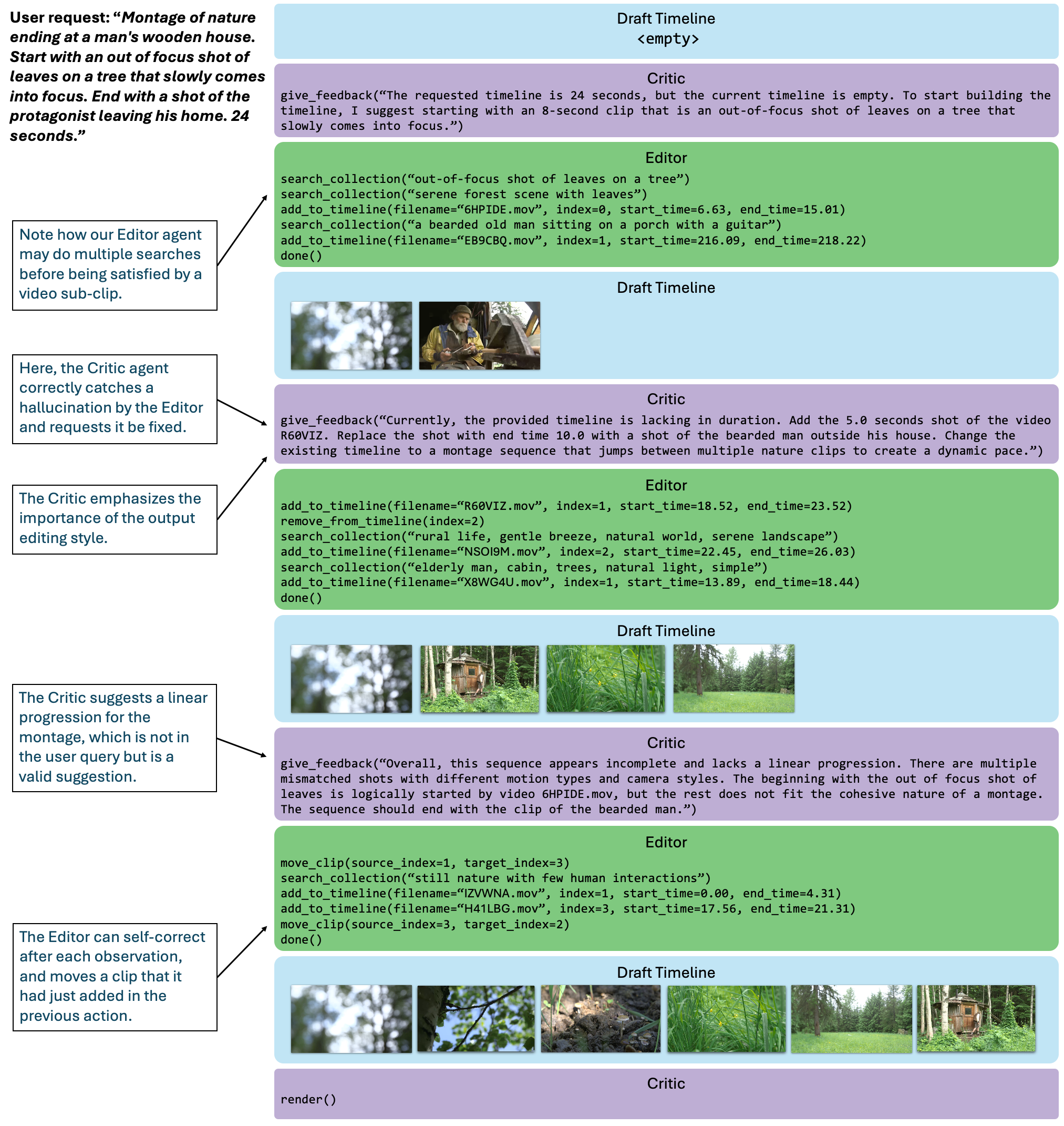}
  \caption{\textbf{Sample run from EditDuet for the documentary ``Built By Life''.} At each iteration, the Critic observes the current state of the timeline, and produces feedback or decides to render it. The Editor may execute any number of actions until it signals to the Critic that it is finished and ready for the next iteration. All thumbnails are copyrighted and belong to EditStock.}\label{fig:qualitative-figure}
\end{figure*}

We provide an example for our agents' interactions in Figure~\ref{fig:qualitative-figure}. It shows some key characteristics of our agents, in particular the importance of specialization in the Critic and the ability of the Editor to self-correct because of step-by-step execution, which is impossible in one-pass systems like VisProg and T2V.

\begin{figure*}
  \includegraphics[height=0.9\textheight]{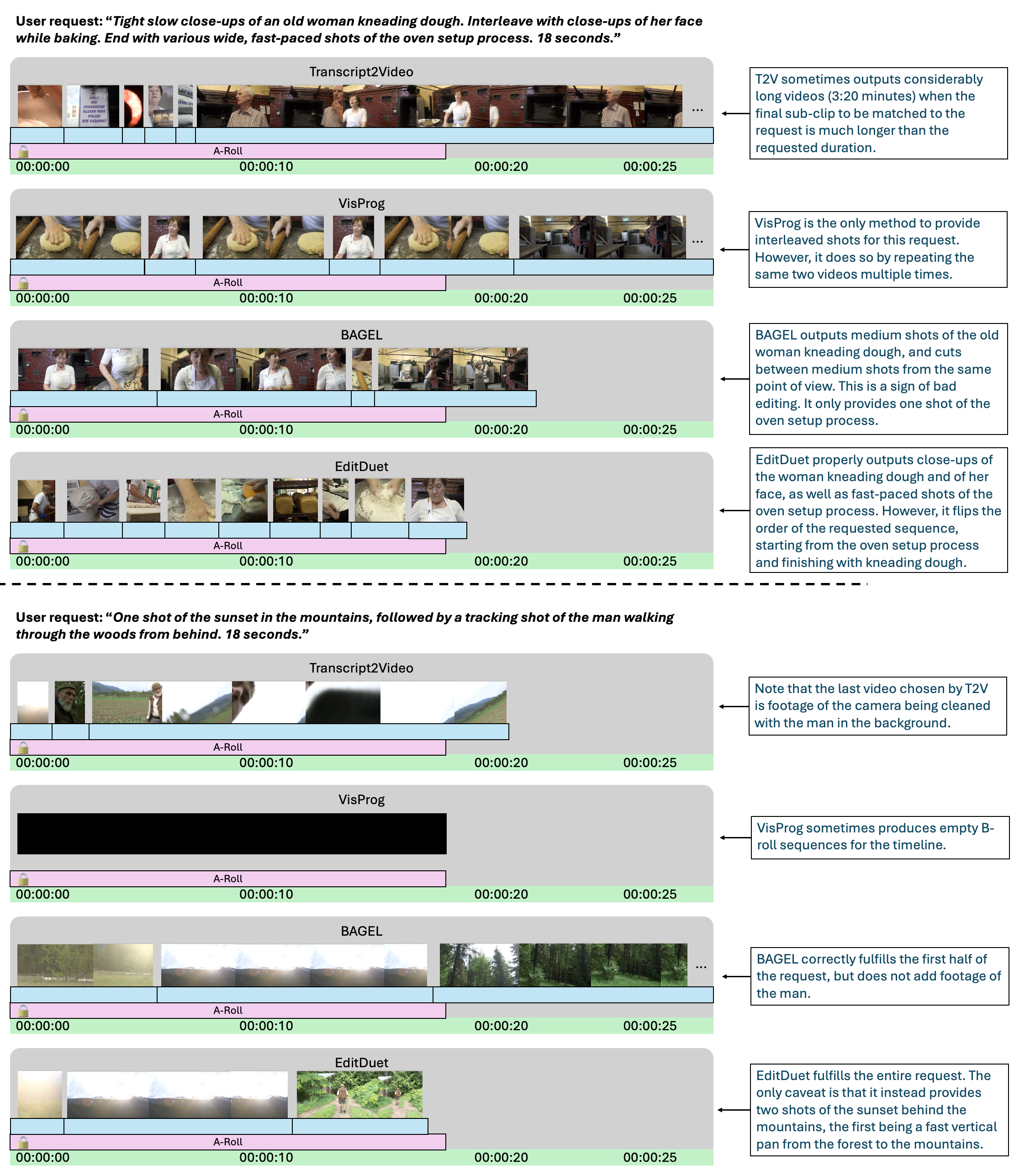}
  \caption{\textbf{Example outputs for Transcript2Video, VisProg, BAGEL, and EditDuet for the documentaries ``Built By Life'' and ``Ovens of Cappoquin''.} Note that our method, EditDuet, is the closest to the desired duration in both samples, as well as the one whose output is semantically closest to the desired request. BAGEL is the second closest, showing the superiority of agentic approaches to video NLE over non-agentic ones (Transcript2Video and VisProg), whose outputs often break or have many unwanted sub-clip repetitions. To watch the specific video outputs, please see the \textit{supplementary material}. All thumbnails are copyrighted and belong to EditStock.}\label{fig:outputs-figure}
\end{figure*}

Figure~\ref{fig:outputs-figure} shows example outputs for two queries from our results for T2V, VisProg, BAGEL, and EditDuet. T2V tends to output footage of low aesthetic quality, for example, a sub-clip where the videographer is cleaning the camera lens. VisProg is prone to repetitions and outputs programs with empty B-roll sequences. Agentic approaches perform better, with BAGEL and EditDuet providing competitive results. Overall, EditDuet produces the B-roll sequences with best quality footage, most appropriate pacing, and closest to the user request.
We hypothesize that the LLMs' ability to understand video editing vocabulary stems from pre-training on Internet-scale data, including editing knowledge from textbooks and tutorials.

\subsection{Limitations}
Exploration helps to greatly reduce, from 19.5\% to 8.2\%, the occurrence system-breaking of failures. These usually take the form of function hallucination, file hallucination, index errors, and adding video sub-clips where the requested start/end times are out of bounds. It also reduces failures that are not system-breaking, like unsupported feedback and nonsensical search queries. These failures occur with and without exploration (Editor Critic vs. EditDuet), but are rarer with exploration.
The most prevalent failure cases in our final system are function hallucination and unsupported feedback.

\section{Conclusion}

In this work, we proposed a multi-agent architecture for non-linear video editing, along with a method for enabling effective communication between multiple agents.
We formulated video NLE as a decision process that is closely aligned to how a human would interact with NLE software, and showed the shortcomings of previous approaches on real-world data.
We showed that EditDuet is able to overcome these, in terms of quantitative metrics, human preference through a user study, and automatic preference judgments.

Future work in this direction includes considerably extending the editing environment in which our agents operate.
Real-world NLE software includes options for creating transitions between sub-clips, edit pixel-level content of videos, or audio editing.
These are crucial video editing operations, and may be addressed by extending the action space of our agents or adding specialized agents for some of these operations.
In addition, as VLMs become more powerful and their vision contexts become larger, it becomes feasible to extend our agents to fully-fledged VLMs capable of processing large amounts of video at once.
Finally, this agents-based approach along with our AI-based evaluation open up possibilities for reinforcement learning-based non-linear video editing agents and models.

\bibliographystyle{ACM-Reference-Format}
\bibliography{main}

\clearpage

\section{Supplementary Material}

\renewcommand{\thesubsection}{\Alph{subsection}}

\subsection{Ethical Considerations}
We believe our system does not represent a risk of automating away the role of human editors, as there are cultural and artistic sensibilities in the process that are not easily captured by any system for automated editing, including ours. However, outputs from such systems can help streamline some of the tedious work associated with video editing and can act as meaningful starting points for human editors' creative processes. Our system interacts in an environment that is compatible with currently used non-linear editing software abstractions, allowing for human input during the editing process and full human control over the final result.

\subsection{Example Failures}
\label{sec:failures}

\begin{figure}[!ht]
  \includegraphics[width=\linewidth]{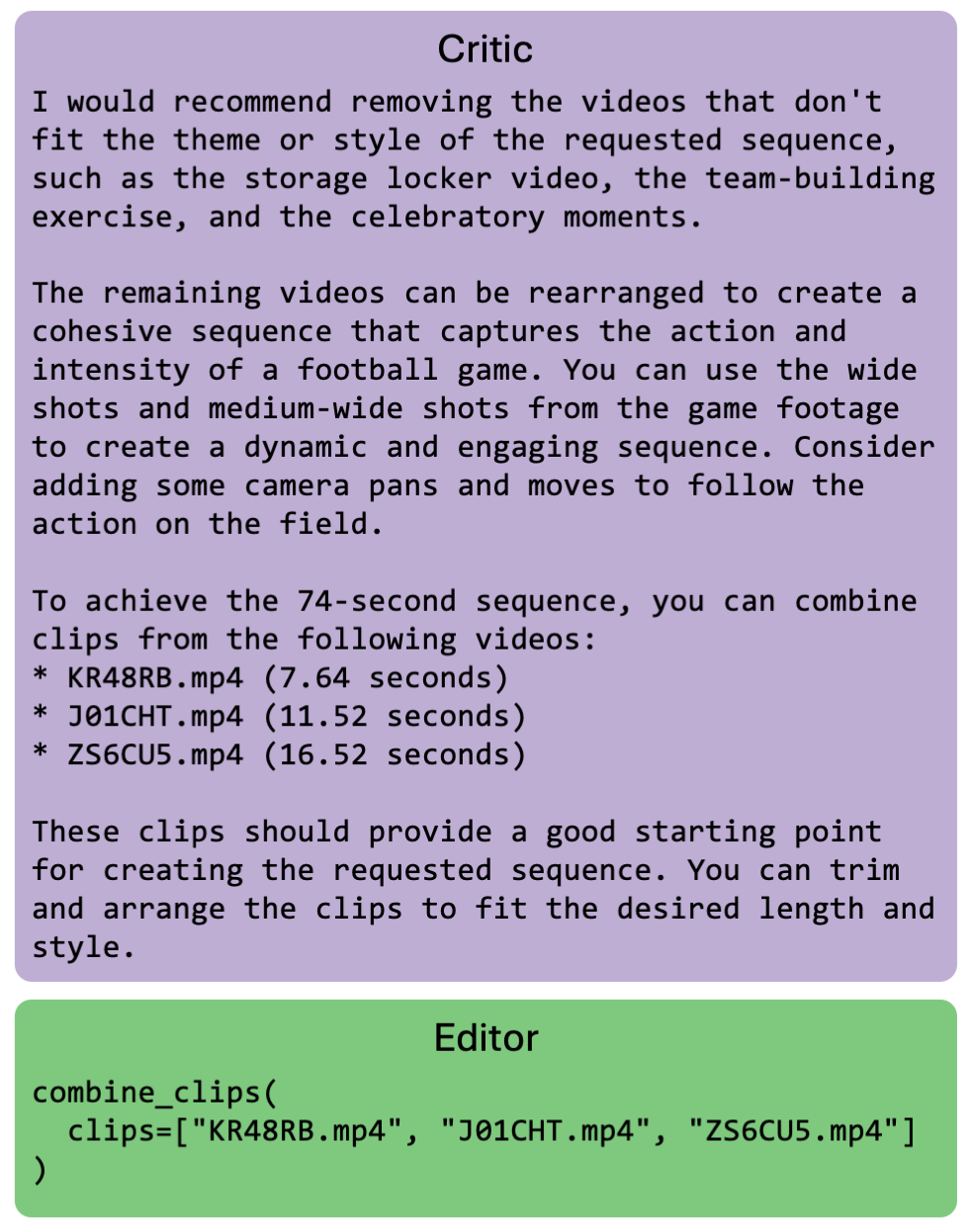}
  \caption{\textbf{Function hallucination.} The Critic suggests combining sub-clips from a few video files (probably meaning that these should be placed in the timeline consecutively), to which the Editor hallucinates a function called \texttt{combine\_clips}. This function does not exist, and thus, the run fails.}
  \label{fig:func-hallucination}
\end{figure}

\begin{figure}[!ht]
  \includegraphics[width=\linewidth]{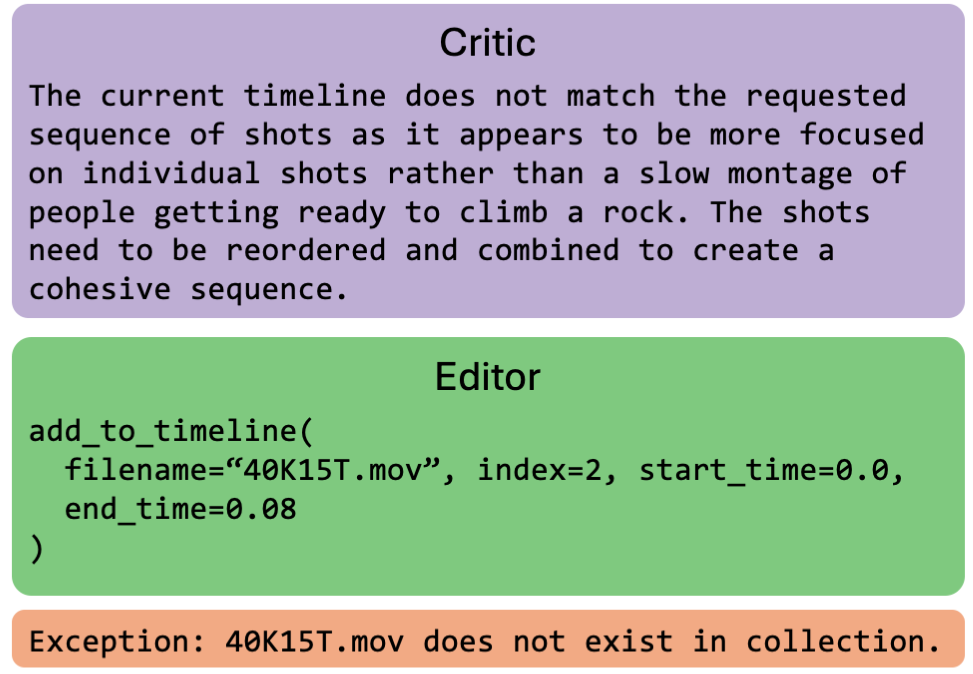}
  \caption{\textbf{File hallucination.} The Editor attempts to add a file to the timeline regardless of what the Critic has suggested. This file does not exist, and thus, the run fails.}\label{fig:file-hallucination}
\end{figure}

\begin{figure}[!ht]
  \includegraphics[width=\linewidth]{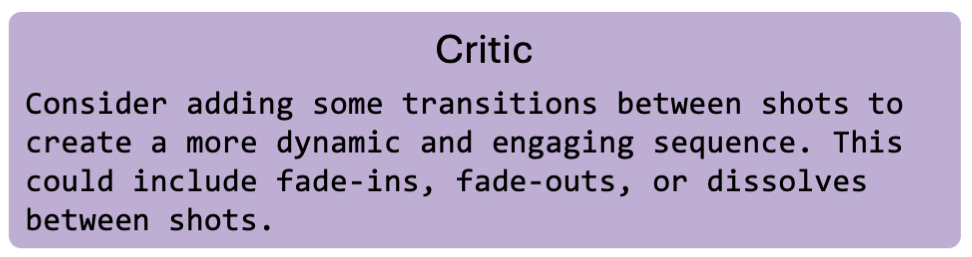}
  \caption{\textbf{Unsupported feedback.} The Critic makes a suggestion that is not possible to fulfill given the current set of tools the Editor has access to, namely, adding transitions.}\label{fig:unsup-feedback}
\end{figure}

\begin{figure}[!ht]
  \includegraphics[width=\linewidth]{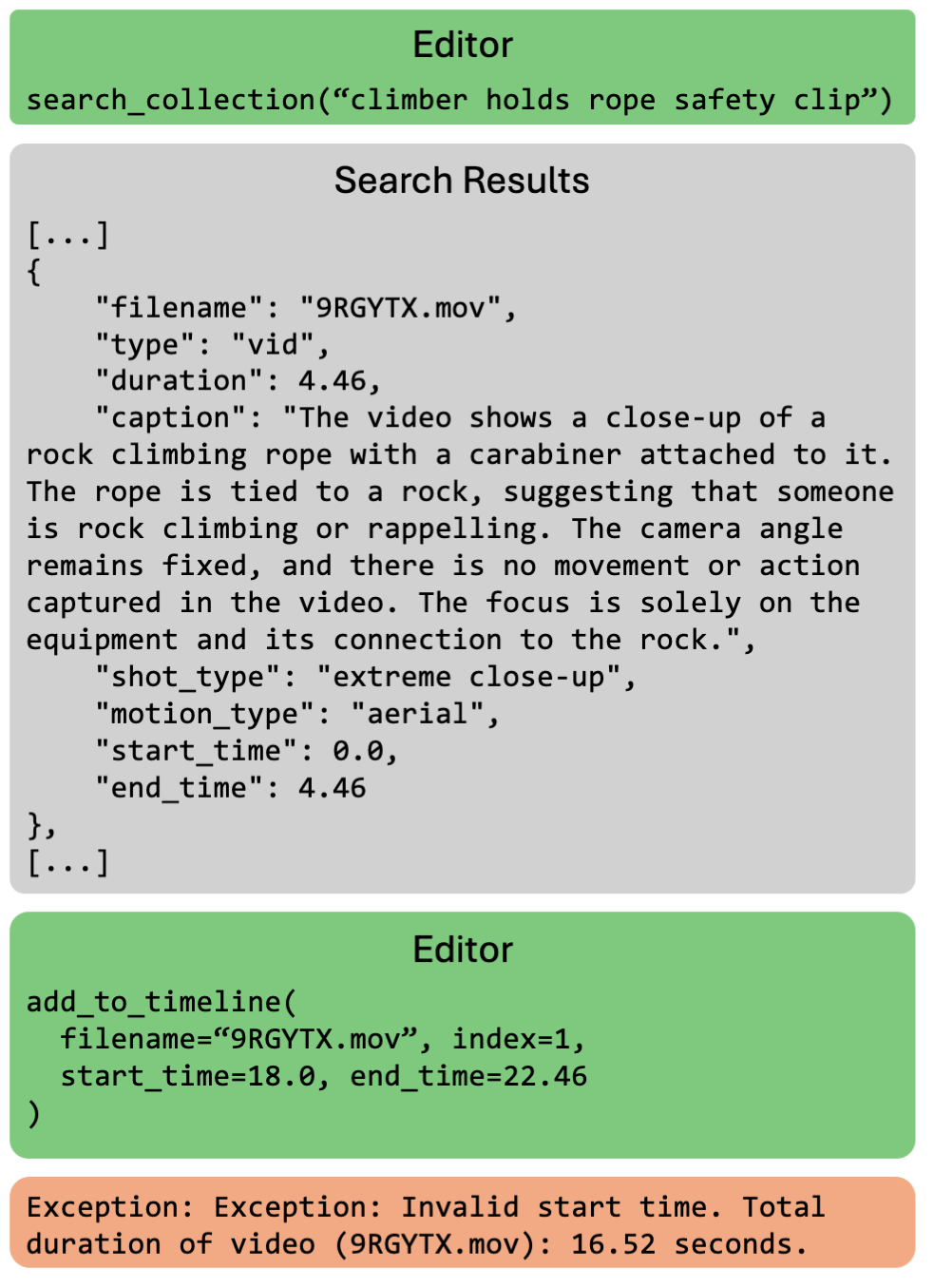}
  \caption{\textbf{Out-of-bounds video sub-clips.} The Editor performs a search in the video collection, and then attempts to add one video sub-clip to the timeline. However, the chosen start time and end time for the sub-clip are both higher than the total duration of the source file, so the run fails.}\label{fig:oob-subclip}
\end{figure}

Figures~\ref{fig:func-hallucination}, \ref{fig:file-hallucination}, \ref{fig:unsup-feedback} and \ref{fig:oob-subclip} show four examples of common failure modes in our Editor Critic system and in our final EditDuet. The most notable ones are function hallucination, file hallucination, unsupported feedback, and out-of-bounds video sub-clips. Other less common examples include index errors and seemingly nonsensical search queries. Our exploration process reduces the occurrence of these significantly.

\subsection{Prompts}
\label{sec:prompts}
\newcounter{prompt}
We include all Prompts (\ref{editor_agent_prompt}, \ref{critic_agent_prompt}, \ref{editor_explorer_prompt}, \ref{editor_labeler_prompt}, \ref{editor_scorer_prompt}, \ref{critic_explorer_prompt}, \ref{critic_labeler_prompt}, \ref{critic_scorer_prompt}) crafted for our method.

\subsection{Baseline Implementation Details}
\noindent\textbf{Transcript2Video.} For CLIP similarity, we use features extracted from the model CLIP-ViT-B-32. As T2V is not originally designed to respect a given duration in seconds, we modify it to do so. Instead of constraining it to number of shots, we keep track of the total duration of the video sequence as we are appending sub-clips, and only finish the process once at least 90\% of the duration has been matched or exceeded.

\noindent\textbf{VisProg.} We use the same LLM backbone as for our agents, Llama3.1-8B-Instruct. Early in our experimentation with VisProg, we found that it often failed at using our standard \texttt{add\_to\_timeline} function, that receives four inputs: the video file name, the index in the timeline, the start time of the sub-clip, and the end time of the sub-clip. To mitigate this, we modified the function to instead be able to take search result objects directly and an index in the timeline.

\subsection{Running Time}
Our exploration method takes approximately 82 minutes. EditDuet inference takes approximately 2.6 minutes on 8 H100s for a single B-roll sequence. T2V and VisProg take less than half a minute on average, and BAGEL takes about 1.5 minutes. High-scoring synthesized demonstrations are, on average, 17 Editor steps long, and 4 Critic steps long.

\subsection{Comparison with Human Editors}
We compared the results of different automatic editing methods to the B-roll sections in the final cuts done by professional editors and provided by EditStock. We use our proposed automatic preference evaluation. The professional version was preferred over T2V (89.7\%), VisProg (92.9\%), BAGEL (96.6\%), and EditDuet (75.1\%). Though these methods are still far away from the professional editors, EditDuet represents a substantial increase in quality compared to other automatic editing methods.

\subsection{Filtered Comparisons}
We also compared all methods' B-roll outputs after filtering repetitions and empty sequences (represented as black frames in our qualitative results). We simply skip these and move back the rest of the video to be one continuous sequence. We find that the performance differences remain similar, with the largest decreases coming from competing methods. T2V's automatic preference goes down to 8.70\% (against our method), VisProg's goes down to 9.09\%, and BAGEL's goes down to 4.35\%. In contrast, the Editor-only method achieves 18.18\% and the Editor Critic achieves 30.77\%. Our approach remains the preferred one in this setting.

\clearpage

\begin{figure*}[!ht]
\noindent
\refstepcounter{prompt}
\begin{tcolorbox}[colback=gray!10, colframe=black!75, width=\textwidth, title=Prompt \#\theprompt: Editor Agent ($\mathcal{E}$)]
\begin{texttt}
You are a video-editing AI trained to interact with text-based editing software. Your task is to satisfy editing requests provided by a user. You have access to a set of functions, a timeline, which will be rendered at the end, a search panel, a collection of videos that you will use to assemble the video requested by the user, and an audio transcription of the voice-over that will be played over your video. Your job is to satisfy the user's query as precisely as possible.\\

Here is a list of tools that you have available:\\
\textbf{\{function definitions in Python\}}\\

\textbf{\{in-context examples\}}
\end{texttt}
\end{tcolorbox}
\label{editor_agent_prompt}
\end{figure*}

\begin{figure*}[!ht]
\noindent
\refstepcounter{prompt}
\begin{tcolorbox}[colback=gray!10, colframe=black!75, width=\textwidth, title=Prompt \#\theprompt: Critic Agent ($\mathcal{C}$)]
\begin{texttt}
You are a video editing critic. You will receive a prompt of what the editing of the video should look like and a shot list with descriptions and durations for each video. A timeline is a list of clips, each of which is represented by a dictionary with that describes a number of features for each clip.\\

Your job is to ensure the provided shot list follows the prompt as closely as possible. Make sure it reasonably respects duration, pacing, content, and coherence.\\

You have two actions available to you, namely `give\_feedback` and `render`.\\

When giving feedback, you will provide the editor some suggestions on how to improve the timeline. Enforcing duration to be close to the request by the user is critical. Ensure your feedback always includes one corrective action. Videos can only be added, removed, and moved. The editor does not see the user request, so you must guide it in achieving this task.\\

On the other hand, taking a `render` action finishes this process and creates a final video. You must only render once you are satisfied with the current timeline. Be very strict and only render when the timeline fully satisfies the request.\\

\textbf{\{in-context examples\}}
\end{texttt}
\end{tcolorbox}
\label{critic_agent_prompt}
\end{figure*}

\begin{figure*}[!ht]
\noindent
\refstepcounter{prompt}
\begin{tcolorbox}[colback=gray!10, colframe=black!75, width=\textwidth, title=Prompt \#\theprompt: Editor Explorer ($\mathcal{E}_{expl}$)]
\begin{texttt}
You are a video-editing AI trained to interact with text-based editing software. Your task is to satisfy editing requests provided by a user. You have access to a set of functions, a timeline, which will be rendered at the end, a search panel, a collection of videos that you will use to assemble a video in the timeline. Your objective is to discover diverse and interesting tasks (that a human might give to an agent) by interacting with the editing software through the tools you are given.\\

Here is a list of tools that you have available:\\
\textbf{\{function definitions in Python\}}\\

\end{texttt}
\end{tcolorbox}
\label{editor_explorer_prompt}
\end{figure*}

\begin{figure*}[!ht]
\noindent
\refstepcounter{prompt}
\begin{tcolorbox}[colback=gray!10, colframe=black!75, width=\textwidth, title=Prompt \#\theprompt: Editor Labeler ($\mathcal{L}_{editor}$)]
\begin{texttt}
A video editing agent interacting with a text-based video editor is given precise feedback from a human, which it carries out through a sequence of sub-tasks, where each sub-task (such as searching for videos, adding them to the timeline, moving around clips in the timeline, and removing videos from the timeline) changes the video to be rendered.\\

Each timeline contains information on the file name, the duration in seconds, and a brief description of the video.\\

You are given the initial timeline and the final timeline after all sub-tasks are carried out.\\

Your objective to guess the feedback paragraph that was given to the agent which led to modifying the original video timeline. Ensure that your feedback is concrete and such that every change in the timeline is reflected in the feedback paragraph. You will first provide your reasoning, which may be of arbitrary length, and then you may summarize the feedback you think the agent received.
\end{texttt}
\end{tcolorbox}
\label{editor_labeler_prompt}
\end{figure*}

\begin{figure*}[!ht]
\noindent
\refstepcounter{prompt}
\begin{tcolorbox}[colback=gray!10, colframe=black!75, width=\textwidth, title=Prompt \#\theprompt: Editor Scorer ($\mathcal{S}_{editor}$)]
\begin{texttt}
A video editing agent interacting with a text-based video editing software is given precise feedback by a human, which it carries out through a sequence of sub-tasks, where each sub-task (such as searching for videos, adding them to the timeline, moving around clips in the timeline, and removing videos from the timeline) changes the video to be rendered.\\

Each timeline contains information on the name of the file, a description of the visual contents of the video, its duration in seconds, the cinematographic shot type most prevalent int he video, and the type of motion observed.\\

You are given the initial timeline, the final timeline after all sub-tasks are carried out, and each intermediate change the sub-tasks made to the starting timeline state. You are also given the original user feedback that led to modifying the original video timeline.\\

Your objective is to provide a score, from 1 to 5, of how well the feedback was incorporated in the actions carried out by the agent. Only give a score of 5 if the changes perfectly reflect the feedback provided by the user and in the fewest possible steps. You will first be asked to provide your reasoning, and only after that is done you will provide your score as a single digit integer.
\end{texttt}
\end{tcolorbox}
\label{editor_scorer_prompt}
\end{figure*}

\begin{figure*}[!ht]
\noindent
\refstepcounter{prompt}
\begin{tcolorbox}[colback=gray!10, colframe=black!75, width=\textwidth, title=Prompt \#\theprompt: Critic Explorer ($\mathcal{C}_{expl}$)]
\begin{texttt}
You are a video editing critic interacting with an editor that carries out any feedback you give it. At the start, you will be provided a timeline that you may edit to your liking.\\

A timeline is a list of clips, each of which is represented by a dictionary that contains information about the visual contents of the video and its duration in seconds.\\

You have two actions available to you, namely `give\_feedback` and `render`.\\

When giving feedback, you will provide the editor some suggestions on how to improve the timeline. Enforcing duration to be close to the request by the user is critical. Ensure your feedback always includes one corrective action. Videos can only be added, removed, and moved. The editor does not see the user request, so you must guide it in achieving this task.\\

On the other hand, taking a `render` action finishes this process and creates a final video. You must only render once you are satisfied with the current timeline.\\

Your objective is to discover diverse and interesting tasks (that a human might give to the agent) by interacting with this editor through your feedback. Remember: the editor is imperfect, and the way you phrase your feedback matters.\\

Possible feedback strings you may give are:\\
\textbf{\{example synthetic labels from Editor exploration\}}\\

Freely interact with the editing agent to construct a timeline to your liking.
\end{texttt}
\end{tcolorbox}
\label{critic_explorer_prompt}
\end{figure*}

\begin{figure*}[!ht]
\noindent
\refstepcounter{prompt}
\begin{tcolorbox}[colback=gray!10, colframe=black!75, width=\textwidth, title=Prompt \#\theprompt: Critic Labeler ($\mathcal{L}_{critic}$)]
\begin{texttt}
You are monitoring two agents interacting with each other in a text-based video editing software. One executes actions (such as searching for videos, adding them to the timeline, moving around clips, and removing videos from the timeline), and the other provides feedback on the resulting timeline or renders it. The task of the feedback-giving agent is to make sure the final timeline is as close as possible to the one requested by the user.\\

Each timeline contains information on the file name, the duration in seconds, and a brief description of the video.\\

You are given the initial timeline, the final timeline after all sub-tasks have been carried out, and each intermediate feedback provided by the feedback giving agent before rendering. At the end of the sequence, the timeline satisfies the user query.\\

User queries are given in terms of editing style, semantic content, and general descriptions of feelings and semiotics. Additionally, the user may have provided a target duration. However, the user has not seen any files in advance.\\

Your objective to guess what the user query was based on the feedback of the agent and the timeline states. Ensure that every piece of feedback meaningfully contributes to an aspect of the user query. Pretend you are the user.
\end{texttt}
\end{tcolorbox}
\label{critic_labeler_prompt}
\end{figure*}

\begin{figure*}[!ht]
\noindent
\refstepcounter{prompt}
\begin{tcolorbox}[colback=gray!10, colframe=black!75, width=\textwidth, title=Prompt \#\theprompt: Critic Scorer ($\mathcal{S}_{critic}$)]
\begin{texttt}
You are monitoring two agents interacting with each other in a text-based video editing software. One executes actions (such as searching for videos, adding them to the timeline, moving around clips, and removing videos from the timeline), and the other provides feedback on the resulting timeline or renders it. The task of the feedback-giving agent is to make sure the final timeline is as close as possible to the one requested by the user.\\

Each timeline contains information on the file name, the duration in seconds, and a brief description of the video.\\

You are given the final timeline the agents produced and the original user query.\\

Your objective is to provide a score, from 1 to 5, measuring how well the user query was fulfilled by the timeline produced by the agents. Only give a score of 5 if the timeline perfectly fulfills what was asked for in the user query.\\
\end{texttt}
\end{tcolorbox}
\label{critic_scorer_prompt}
\end{figure*}

\end{document}